\documentclass[final]{cvpr}

\usepackage{times}
\usepackage{epsfig}
\usepackage{graphicx}
\usepackage{amsmath}
\usepackage{amssymb}
\usepackage{enumitem}

\usepackage{multirow}
\usepackage{color}
\definecolor{gray}{rgb}{0.5,0.5,0.5} 
\definecolor{ao}{rgb}{0.0, 0.5, 0.0}

\usepackage{diagbox}
\usepackage{colortbl}
\definecolor{byzantine}{rgb}{0.74, 0.2, 0.64}

\usepackage[colorlinks]{hyperref}

\def\cvprPaperID{573} 
\def\confYear{CVPR 2021}

\begin{document}
\setlength{\abovedisplayskip}{3pt}
\setlength{\belowdisplayskip}{3pt}
\title{A3D: Adaptive 3D Networks for Video Action Recognition}

\author{Sijie Zhu\thanks{Equal contribution}, \hspace{0.25em} Taojiannan Yang\footnotemark[1], \hspace{0.25em} Matias Mendieta, \hspace{0.2em} Chen Chen\\
Department of Electrical and Computer Engineering, University of North Carolina at Charlotte\\
{\tt\small \{szhu3, tyang30, mmendiet, chen.chen\}@uncc.edu}
}

\maketitle

\begin{abstract}
This paper presents A3D, an adaptive 3D network that can infer at a wide range of computational constraints with one-time training. Instead of training multiple models in a grid-search manner, it generates good configurations by trading off between network width and spatio-temporal resolution. Furthermore, the computation cost can be adapted after the model is deployed to meet variable constraints, for example, on edge devices. Even under the same computational constraints, the performance of our adaptive networks can be significantly boosted over the baseline counterparts by the mutual training along three dimensions. When a multiple pathway framework, e.g. SlowFast, is adopted, our adaptive method encourages a better trade-off between pathways than manual designs. Extensive experiments on the Kinetics dataset show the effectiveness of the proposed framework. The performance gain is also verified to transfer well between datasets and tasks. Code will be made available.
\end{abstract}

\section{Introduction}
Spatio-temporal (3D) networks have achieved excellent performance on video action recognition \cite{c3d,i3d,slowfast,x3d,r2+1d,multifiber}, as well as its downstream tasks \cite{actiondet1,actiondet2,actiondet3}. However, popular 3D networks \cite{c3d,i3d,s3d} are extremely computational demanding, with tens or even hundreds of GFLOPs per video clip. Such a large computational cost imposes limits on its applicability in real-world scenarios, especially on resource-limited edge devices. A thread of works \cite{slowfast,x3d,multifiber,wu2019adaframe} has been proposed to reduce the computation cost of 3D networks to meet lower computational budgets. These works narrow the gap between 3D networks and their applications to some degree, but they all ignore the crucial fact that the computational budgets change in many real applications. 
For example, the battery condition of mobile devices imposes constrains on the computational budget of many operations.
Similarly, a task may have specific priorities at any given time, requiring a dynamic computational budget throughout its deployment phases.
To meet different resource constraints, one needs to deploy several 3D networks on the device and switch among them. This consumes much larger memory footprints than a single model and the cost of loading a different model is not negligible. 

In this paper, we propose adaptive 3D networks (A3D) where one model can meet a variety of resource constraints. The computational cost of a 3D network is determined by the input size (spatial and temporal) and network size (width and depth).
During training, we randomly sample several spatial-temporal-width configurations. In this way, the network can perform inference with many configurations in real deployment, making it possible to meet a wide range of computational budgets. Besides, distinct network configurations can learn different semantic information. For example, a larger spatial size can capture more fine-grained features, and a longer temporal duration can encode more long-term semantics. Motivated by this finding, in each training iteration, we randomly sample several configurations and train them jointly. Moreover, we propose Spatial-Temporal Distillation (STD) to transfer the knowledge in larger spatial-temporal configurations to other configurations. This allows smaller configurations to learn fine-grained and long-term representations with less computational cost. As a result, the performance of every configuration is greatly improved. 

Our work shares a similar motivation with X3D \cite{x3d} -- both leverage different network configurations for various computational budgets. However, X3D expands a 2D network along different dimensions. In each expansion step, X3D trains 6 models which correspond to 6 dimensions. To obtain a moderate size model (\eg X3D-L), it requires 10 steps, which means that it needs to train 60 models in the whole process. This makes the expansion very inefficient. However, in A3D, we train different configurations jointly. Therefore, the whole framework is end-to-end, making it simpler and more efficient than X3D. Besides, during inference, X3D has to deploy several different scaled models to meet dynamic computational budgets. On the other hand, A3D only needs to deploy one model, making it abundantly more memory-friendly on edge devices. Furthermore, X3D trains each configuration independently, which fails to leverage the variety of semantic information contained in different configurations. 
However, in A3D, each configuration can transfer its knowledge to one another through the proposed Spatial-Temporal Distillation; this enables every configuration to learn better representations.

Our experimental results reveal that A3D outperforms independently trained models under the same network configuration, even when that configuration is not the best performing one A3D found at that computational budget. Note that A3D can also be applied to the network structures in X3D. We conclude our contributions as follows:
\setlist{nolistsep}
\begin{itemize}[noitemsep,leftmargin=*]
    \item We are the first to achieve adaptive 3D networks where one model can meet different computational budgets in real application scenarios. For training, the proposed A3D requires just a fraction of the training cost compared with independently training several models. For inference, A3D deploys only one model rather than several independently trained models to cope with dynamic budgets.
    \item The proposed Spatial-Temporal Distillation (STD) scheme transfers knowledge among different configurations. This allows every configuration to learn multi-scale spatial and temporal information, thereby improving the overall performance.
    \item On Kinetics-400, A3D outperforms its independently trained counterparts at various computational budgets, especially when the budget is low. The effectiveness of the learned representation is also validated via cross-dataset transfer (Charade dataset \cite{charade}) and cross-task transfer (action detection on AVA dataset \cite{gu2018ava}).
\end{itemize}


\section{Related Work}
\noindent \textbf{Spatio-temporal (3D) Networks.}
The basic idea of video recognition architectures stems from 2D image classification models. \cite{c3d, i3d, 3dresnet, s3d} build 3D networks by extending 2D convolutional filters \cite{vgg, googlenet, resnet, resnext, densenet} to 3D filters along the temporal axis; then the 3D filters can learn spatio-temporal representations in a similar way to their 2D counterparts. Later works \cite{s3d, p3d, r2+1d, slowfast} propose to treat the spatial and temporal domains differently. \cite{s3d} reveals that a bottom-heavy structure is better than naive 3D structures in both accuracy and speed. \cite{p3d, r2+1d} propose to split 3D filters to 2D+1D filters, which reduce the heavy computational cost of 3D filters and improve the performance. SlowFast \cite{slowfast} further shows that space and time should not be handled symmetrically, and introduces a two-path structure to deal with slow and fast motion separately. Recently, \cite{nasvideo1, nasvideo2, nasvideo3} explore neural architecture search (NAS) techniques to automatically learn spatio-temporal network structures.

\noindent \textbf{Efficient 3D Networks.}
3D networks are often very computationally expensive. Many approaches \cite{r2+1d, p3d, s3d, x3d, slowfast, multifiber, TSM, OCT, groupedspatiotemporal, CSN} have been proposed to reduce the complexity. \cite{r2+1d, p3d, s3d} share the idea of splitting the 3D filters to 2D and 1D filters. \cite{multifiber, groupedspatiotemporal, CSN} leverage the group convolution and channel-wise separable convolution in 2D networks \cite{howard2017mobilenetsv1, sandler2018mobilenetv2} to reduce computational cost. Other approaches such as \cite{OCT, TSM} propose generic, insertable modules to improve temporal representations with minimal computational overhead. \cite{x3d, slowfast} improve efficiency by performing trade-offs across several model dimensions. A few methods \cite{wu2019adaframe, arnet, korbar2019scsampler} also explore adaptive frame sampling to reduce computations. However, 
\emph{none of these methods can achieve dynamic computational budgets} and our approach is complementary to these light-weight structures.

\noindent \textbf{Multi-dimension Networks.}
The computational cost and accuracy of a model is determined by both the input size and network size. In 2D networks, there is a growing interest \cite{yang2020mutualnet, yu2020bignas, onceforall, tan2019efficientnet, multidimpruning} in achieving better accuracy-efficiency trade-offs by balancing different model dimensions (\eg image resolution, network width and depth). EfficientNet \cite{tan2019efficientnet} performs a grid-search on different model dimensions and expands the configuration to larger models. \cite{yang2020mutualnet, yu2020bignas, onceforall} train different configurations jointly so that one model can perform variable execution in the 2D domain. \cite{multidimpruning} prunes networks from multiple dimensions to achieve better accuracy-complexity trade-offs. X3D \cite{x3d} is the first work to investigate the effect of different dimensions in spatio-temporal networks. It expands a 2D network step by step to a 3D one. Our method also aims to achieve better model configurations under different budgets. However, we train various intrinsic configurations and share knowledge between them, creating a \emph{time-saving end-to-end framework that enables effective adaptive inference with a single model.}

\section{Adaptive 3D Network}
\begin{figure*}[!htbp]
    \centering
    \vspace{-0.2cm}
    \includegraphics[width=0.97\linewidth]{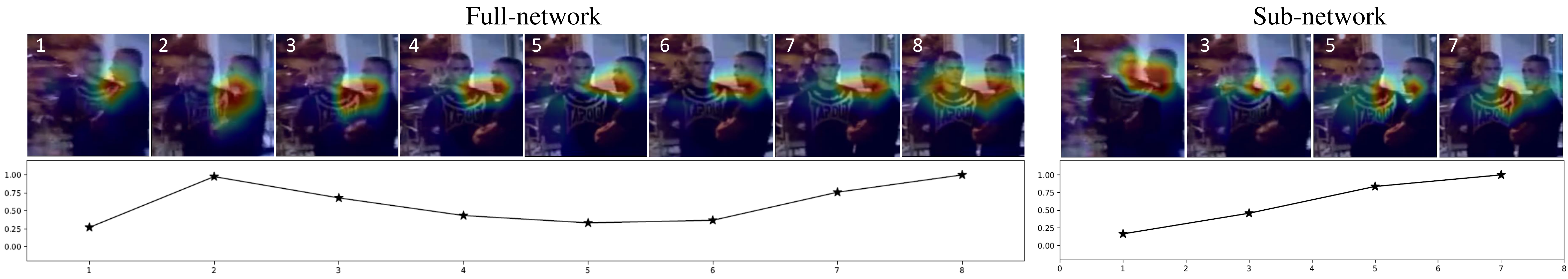}
    \vspace{-0.1cm}
    \caption{Class activation maps along spatial and temporal dimensions of the full-network and sub-network.}
    \label{fig:slow_cam}
\end{figure*}
\begin{figure*}[!htbp]
    \centering
    \includegraphics[width=0.97\linewidth]{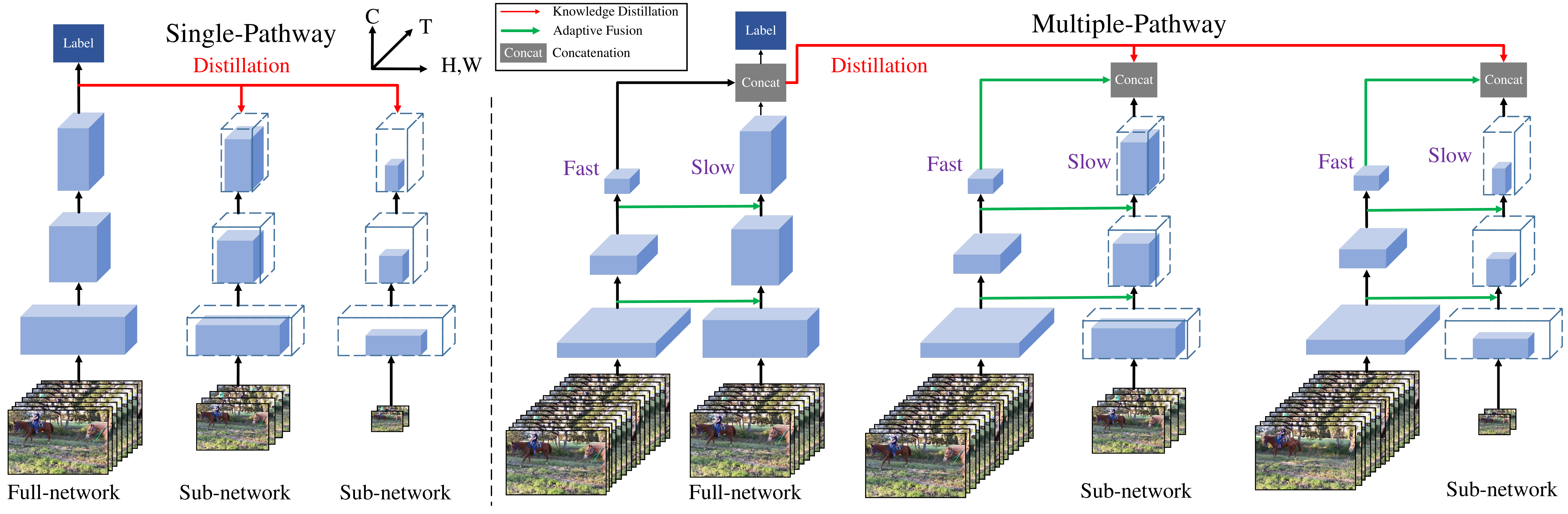}
    \vspace{-0.1cm}
    \caption{An overview of the spatial-temporal distillation strategy to facilitate knowledge transfer among different network configurations.}
    \label{fig:framework}
\end{figure*}
Standard 3D models are trained at a fixed spatial-temporal-width configuration (\eg 224-8-1.0$\times$). But the model can not be well generalized to other configurations during inference. In A3D, we randomly sample different spatial-temporal-width configurations during training, so the model can run at different configurations during inference. Note that the computation cost of a vanilla 3D convolutional layer is given by
\begin{equation}
\label{eq:1}
    K\times K \times C_{i}\times C_{o} \times H \times W \times T.
\end{equation}
Here, $K$ denotes the kernel size, and $C_{i}$, $C_{o}$ are the input and output channels of this layer. $H$, $W$, $T$ denote the spatial-temporal size of the output feature map. For a sub-network with network width (channels) coefficient $\gamma_{w} \in [0,1]$, and spatial-temporal resolution  factors $\gamma_{s},\gamma_{t} \in [0,1]$, the computation cost is reduced to
\begin{equation}\label{eq:2}
    K\times K \times \gamma_{w}C_{i}\times \gamma_{w}C_{o} \times \gamma_{s}H \times \gamma_{s}W \times \gamma_{t}T.
\end{equation}
The computational cost is now $\gamma_{w}^{2}\gamma_{s}^{2}\gamma_{t}$ times that of the original in Eq. \ref{eq:1}. Our goal is to train a 3D network that is executable at a range of resource constraints.
For example, we define a maximal reduction coefficient $\rho$ in which $\gamma_{w}^{2}\gamma_{s}^{2}\gamma_{t} \in [1/\rho, 1.0]$. This allows for an execution range from a $\rho$ times reduction in computation up to the full network. With this range, one possible configuration set lets $\gamma_{w}, \gamma_{s} \in [\sqrt[\leftroot{-1}\uproot{3}6]{1/\rho}, 1.0]$ and $\gamma_{t} \in [\sqrt[\leftroot{-1}\uproot{3}3]{1/\rho}, 1.0]$. Here, each dimension is equally responsible for a $\sqrt[\leftroot{0}\uproot{3}3]{\rho}$ times computation reduction. Further details will be discussed in the following sections.

In Sec. \ref{sec:STD}, we first show that different model configurations focus on different semantic information in a video. Then we demonstrate how the knowledge is transferred between different sub-networks for single-pathway models. Sec. \ref{sec:multiple} further illustrates the multiple-pathway trade-off when using A3D with SlowFast-inspired networks \cite{slowfast}.

\subsection{Spatial-Temporal Distillation}
\label{sec:STD}
\noindent\textbf{Knowledge in Different Configurations.} Since we randomly sample various model configurations in each training iteration, we want these configurations to learn from each other. However, is there any unique knowledge in a given sub-network that is beneficial for transferring to others? The answer is yes when sub-networks are fed with different spatial-temporal resolutions. Fig. \ref{fig:slow_cam} shows the spatial and temporal distributions of network activation following the spirit of \cite{cam}. Higher value means more contribution to the final logit value. Although both the full-network ($\gamma_{w}=1$) and sub-network generate the prediction as ``headbutting", their decisions are based on different areas of the frames. The input of the full-network has $8$ frames, and the 2nd and 8th frames contribute the most to the final prediction. Since these two key frames are not sampled in the input of the sub-network, it has to learn other semantic information, forcing a change in both temporal and spatial activation distributions. For example, the activation value of the 5th frame exceeds that of the 3rd frame in the sub-network. The attention areas are also unique between networks, indicating that a varied set of visual cues is captured.


\noindent\textbf{Mutual Training.} To fully leverage the semantic information captured by different model configurations, we propose the mutual training scheme as shown in Fig. \ref{fig:framework}. The left half of Fig. \ref{fig:framework} shows how mutual training works in single pathway structures. In each training iteration, we randomly sample two sub-networks (by the width factor $\gamma_w$) in addition to the full-network. Sub-networks share the parameters with part of the full-network. For example, the sub-network with $\gamma_{w}=0.5$ shares the first half of the full-network's parameters in each layer. During training, since the full-network has the best learning capacity, it is always fed with the highest spatial-temporal resolution ($\gamma_{s}=\gamma_{t}=1.0$) inputs. For these two sub-networks, one is randomly sampled with $\gamma_{w} \in [0.5,1.0]$, and the other is always sampled with the minimal $\gamma_{w}=0.5$. The spatial-temporal resolution of a sub-network's input is randomly sampled with $\gamma_{s} \in \{0.57, 0.71, 0.86, 1.0\}$ and $\gamma_{t} \in \{0.25, 0.5, 0.75, 1.0\}$. This allows sub-networks to learn different semantic information as motivated in Fig. \ref{fig:slow_cam}. Since sub-networks share the weights with part of the full-network, the full-network can also benefit from these diverse information. The full-network is trained with ground-truth labels with the Cross Entropy (CE) loss, while the outputs of the full-network are adopted as soft labels for sub-networks using a Kullback-Leibler (KL) Divergence loss. This forces the full-network and sub-networks to have a high-level semantic consistency, even though they focus on different parts of the input. The overall loss function can be written as
\begin{equation}
\begin{split}
    \mathcal{L}& = \underbrace{CE	\Big(\mathcal{F}(I_{\gamma_{s},\gamma_{t}=1}, \Theta_{\gamma_{w}=1}), Y\Big)}_{\textrm{Full-network}}\\
    &+ \underbrace{\sum KL\Big(\mathcal{F}(I_{\gamma_{s},\gamma_{t}=1}, \Theta_{\gamma_{w}=1}),\mathcal{F}(I_{\gamma_{s},\gamma_{t}}, \Theta_{\gamma_{w}})\Big)}_{\textrm{Sub-networks}}.
\end{split}
\end{equation}
$\mathcal{F}$ denotes the network architecture and $I_{\gamma_{s},\gamma_{t}}$ is the input tensor. $\Theta_{\gamma_{w}}$ denotes the parameters of the sub-network and $Y$ is the one-hot class label. 

\subsection{Multiple-Pathway Trade-off}
\label{sec:multiple}
It may be naive to simply treat time as a dimension of the input matrix, as slow and fast motions contain different information for identifying an action class. SlowFast \cite{slowfast} shows that a lightweight fast pathway is a good complement to slow networks. This inspires us to leverage multiple-pathway trade-offs in our A3D framework. The structure is shown in the right half of Fig. \ref{fig:framework}. Since the fast pathway is lightweight (about $10\%$ of the overall computation), reducing its spatial-temporal resolution or network width is not beneficial for the overall computation-accuracy trade-off. In multiple-pathway A3D, we keep the respective $\gamma_{w},\gamma_{s},\gamma_{t}=1$ for the Fast pathway, so that it can provide complementary information for the Slow path with its own $\gamma_{w},\gamma_{s},\gamma_{t}\leq 1$. \emph{Note that this complementary information is not only on temporal resolution but also on spatial resolution.} Furthermore, the multiple-pathway A3D enables a better trade-off on network width than manual design. \\
\noindent\textbf{Adaptive Fusion.}
Given fixed temporal resolutions for two pathways, the fusion is conducted by lateral connections with time-strided convolution in SlowFast \cite{slowfast}.  \emph{However, since all the three dimensions ($\gamma_{w}$, $\gamma_{s}$, $\gamma_{t}$) of the Slow pathway can change during training, directly applying the time-strided convolution does not work for our framework.} Therefore, we design an adaptive fusion block for multiple-pathway A3D.
\begin{figure}[!htbp]
    \centering
    \vspace{-0.1cm}
    \includegraphics[width=0.7\linewidth]{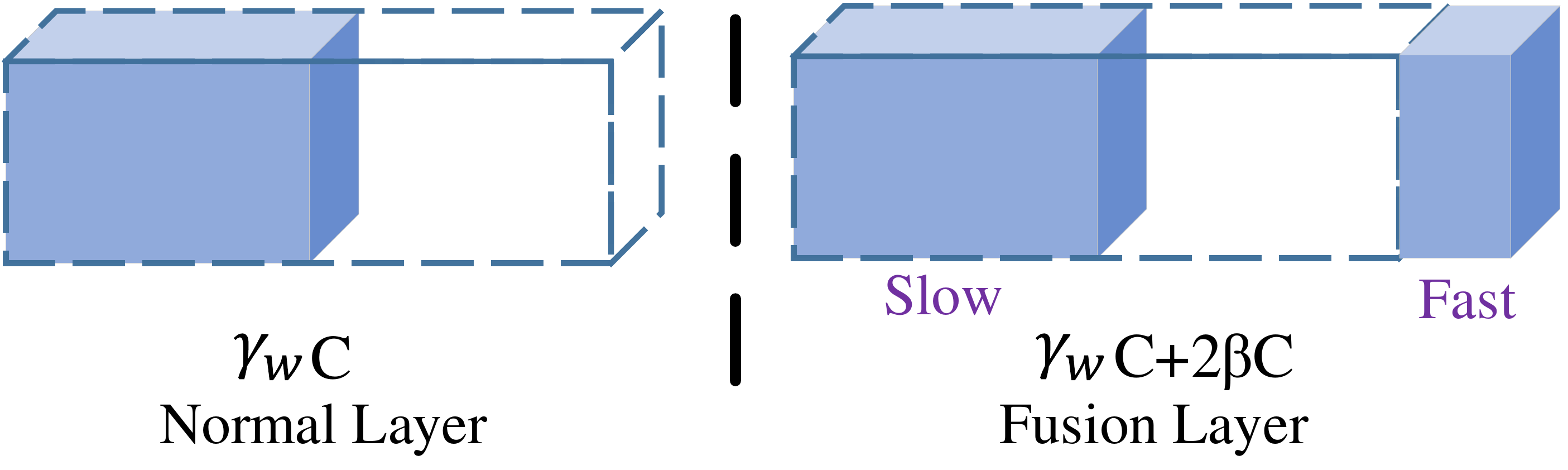}
    \vspace{-0.1cm}
    \caption{The adaptive fusion on network channels.}
    \label{fig:fusion}
    \vspace{-0.1cm}
\end{figure}

Following SlowFast \cite{slowfast}, we denote the feature shape of a standard Slow pathway as $\{T,S^{2},C\}$, where $S$ is the spatial resolution and $C$ is the channel number. Then the feature shape of adaptive Slow pathway is $\{\gamma_{t}T,(\gamma_{s}S)^{2},\gamma_{w}C\}$. The feature shape of Fast pathway remains $\{\alpha T, S^{2},\beta C\}$ as in \cite{slowfast} ($\alpha=8,\beta=1/8$ for SlowFast 4$\times$16). We first perform a 3D convolution of $5\times 1^{2}$ kernel with $2\beta C$ output channels and a stride of $\alpha$. The output feature shape of this convolution layer is $\{T, S^{2}, 2\beta C\}$. To fuse it with the adaptive Slow pathway, we perform a spatial interpolation and temporal down-sampling to make the output shape $\{\gamma_{t}T,(\gamma_{s}S)^{2},2\beta C\}$. Then the final feature shape after the fusion is $\{\gamma_{t}T,(\gamma_{s}S)^{2},(\gamma_{w}+2\beta)C\}$. As shown in Fig. \ref{fig:fusion}, normal convolution layers of adaptive Slow pathway have $\gamma_{w}C$ channels indexing from the left, while the first convolution layer after each fusion has $\gamma_{w}C+2\beta C$ input channels. The last $2\beta C$ channels are always kept for the output of Fast pathway, while the first $\gamma_{w}C$ channels from the left can vary for each iteration. This operation enforces an exact channel-wise correspondence between the fusion features and the parameters in convolution layers.


\subsection{Model Inference}
After training, A3D models can run at different model configurations. We test the performance of different configurations on the validation set. Then we choose the best performing one under each resource budget. For example, we can train a single-pathway A3D model with a Slow \cite{slowfast} network backbone, notated as A3D-Slow-8$\times$8-[0.016,1.0]. Here, 8$\times$8 indicates the number of input frames and temporal stride respectively, and [0.016, 1.0] is the adaptive computation range for $\gamma_{w}^{2}\gamma_{s}^{2}\gamma_{t}$ with $\rho = 64$. Then, we test configurations within this computational range, where $\gamma_{w} \in \{0.5, 0.6, 0.7, 0.8, 0.9, 1.0\}$, $\gamma_{s} \in \{0.57, 0.71, 0.86, 1.0\}$, $\gamma_{t} \in \{0.25, 0.5, 0.75, 1.0\}$. This provides a total of 96 configurations.  After that we choose the best configuration at certain computation budgets to form the configuration-budget table. One can adjust the values of $\gamma_w, \gamma_s$ and $\gamma_t$ to get a more fine-grained table. Note that \textbf{this process is only done once}, so it is very efficient. For real deployment, we only need to keep the model and the table, and therefore the memory consumption is essentially the same as a single model. Given a resource constraint, we can adjust the model according to the configuration-budget table. During inference, we perform batch normalization (BN) calibration as proposed in \cite{usnet} since the BN means and variances are distinct in different configurations. \emph{We do not perform any re-training during inference for all configurations.}

\section{Experiments}
We use SlowFast \cite{slowfast} as our backbone network since it is simple, clean and has the state-of-the-art performance. We conduct experiments on three video datasets following the standard evaluation protocols. We first evaluate our method on Kinetics-400 \cite{kay2017kinetics} for action classification. Then we transfer the learned representations to Charade \cite{charade} action classification and AVA \cite{gu2018ava} action detection. Finally, we perform extensive ablation studies to analyze the effect of different components in A3D.

\noindent\textbf{Datasets.} Kinetics-400 \cite{kay2017kinetics} is a large scale action classification dataset with $\sim$240k training videos and 20k validation videos trimmed as 10s clips. However, since some of the YouTube video links have expired, we can not download the full dataset. Our Kinectic-400 only has 237,644 out of 246,535 training videos and 19,761 validation videos. The training set is about $4\%$ less than that in SlowFast \cite{slowfast} and some of the videos have a duration less than 10s. This leads to an accuracy drop of 0.6\% on Slow-8$\times$8 and 1.2\% on SlowFast-4$\times$16 as we reproduce the results with the officially released codes \cite{codes}.

Charade \cite{charade} is a multi-label action classification dataset with longer activity duration. The average activity duration is $\sim$30 seconds. The dataset is composed of $\sim$9.8k training videos and 1.8k validation videos in 157 classes. The evaluation metric is mean Average Precision (mAP).

AVA \cite{gu2018ava} is a video dataset for spatio-temporal localization of human actions. It consists of 211k training and 57k validation video segments. We follow previous works \cite{slowfast} to report the mean Average Precision (mAP) on 60 classes using an IoU threshold of 0.5.

\noindent\textbf{Training.}
For single-pathway structures, we adopt Slow 8$\times$8 as our backbone. While for multiple-pathway structures we use SlowFast 4$\times$16 due to the limit of GPU memory. For A3D-Slow networks, we train the model for two dynamic budget ranges, [0.06, 1.0] and [0.016, 1.0]. Accordingly, for the range of [0.016, 1.0], the width factor $\gamma_{w}$ is uniformly sampled from $[0.5, 1.0]$. The spatial resolution factor is $\gamma_{s} \in \{0.57, 0.71, 0.86, 1.0\}$ (corresponding to $\{128, 160, 192, 224\}$) and the temporal resolution factor is $\gamma_{t} \in \{0.25,0.5,0.75,1.0\}$ (corresponding to $\{2,4,6,8\}$). For the range of [0.06, 1.0], the width factor $\gamma_{w}$ is uniformly sampled from $[0.63, 1.0]$. The spatial resolution factor is $\gamma_{s} \in \{0.63, 0.80, 1.0\}$ (corresponding to $\{142, 178, 224\}$) and the temporal resolution factor is $\gamma_{t} \in \{0.4, 0.63, 1.0\}$ (corresponding to $\{3, 5, 8\}$). The A3D-SlowFast is only trained with the setting of $[0.06, 1]$. All the models are based on ResNet-50 (R-50) if not specified. Other training settings are the same as the official SlowFast codes \cite{codes}. On Charades and AVA datasets, we finetune the model trained on Kinetics-400 following the same setting as SlowFast \cite{codes}.


\subsection{Main Results}
\noindent\textbf{Kinetics.} In Table \ref{tab:main_result}, we show the classification accuracy and computation cost of the proposed A3D and previous works. We use the Slow network and SlowFast \cite{slowfast} as our baselines. ``-R'' and ``-P'' denote our reproduced results with the official codes and numbers in SlowFast paper, respectively. Given the same implementation environment and dataset, the proposed A3D constantly outperforms its baseline counterpart. And it is executable for a range of computation constraints, denoted as ``model name$\times $ ratio'' in Table \ref{tab:main_result}, \eg a model with $\times 0.5$ only has $50\%$ of the computation of the $\times 1$ model. Although our reproduced accuracy is lower than the reported numbers in SlowFast paper \cite{slowfast} due to the dataset issue, our results are still comparable or better than previous works under the same computation budgets.

Recently, X3D \cite{x3d} achieves the state-of-the-art accuracy with extremely low computation cost (GFLOPs). We believe that using X3D as a baseline in the A3D framework could improve the accuracy-computation trade-off. However, we are not able to perform such experiments due to three reasons: 1) X3D uses 3D depth-wise convolution, which is not well supported in current deep learning platform and GPU acceleration library. Although the GFLOPs of X3D is much smaller than Slow networks \cite{slowfast}, the training time is actually three times more than that of Slow networks. 2) The depth-wise convolution requires high memory access cost (MAC) \cite{ma2018shufflenet}, which further limits its usability in practice. 3) X3D leverages Squeeze-Excitation (SE)-block \cite{senet} and Swish activation \cite{swish} in the network design, which boost the Kinetics accuracy by $1.6\%$ and $0.9\%$, respectively. It is unsuitable to adopt X3D as a baseline to compare with other clean architectures, \eg SlowFast \cite{slowfast}.
\begin{table}[!htbp]
\Large
    \centering
    \resizebox{\linewidth}{!}{
    \begin{tabular}{l|c|c|c|c|c|c}
        model & pre& flow & top-1 & top-5 & GFLOPs$\times$views& Param\\
        \hline
        
        \hline
        I3D~\cite{i3d} & \multirow{7}{0.3cm}{\rotatebox{90}{ImageNet}}& & 71.1 & 90.3 & 108$\times$N/A& 12M\\
        Two-Stream I3D~\cite{i3d} & ~ &\checkmark & 75.7& 92.0 & 216$\times$N/A & 25M\\
        Two-Stream S3D-G \cite{s3d} &~&\checkmark& 77.2 & 93.0 & 143$\times$N/A & 23.1M\\
        MF-Net \cite{multifiber} & ~ & & 72.8 & 90.4 & 11.1$\times$50 &8.0M \\
        TSM R50 \cite{TSM} & ~& & 74.7 & N/A& 65$\times$10 &24.3M\\
        Nonlocal R50 \cite{nonlocal} &~& & 76.5 & 92.6 & 359$\times$30 & 54.3M\\
        \hline
        Two-Stream I3D \cite{i3d} & - & \checkmark&71.6 & 90.0 & 216$\times$NA & 25.0M \\
        R(2+1)D \cite{r2+1d} & - &  & 72.0 & 90.0 & 152$\times$115 & 63.6M\\
        Two-Stream R(2+1)D \cite{r2+1d} & - & \checkmark & 73.9 & 90.9 & 304$\times$115 & 127.2M\\
        ip-CSN-152 \cite{CSN} & - & & 77.8 & 92.8 & 109$\times$30 & 32.8M \\
        \color{gray}{X3D-M-P \cite{x3d}}&-& ~& \color{gray}{76.0} & \color{gray}{92.3} & \color{gray}{6.2$\times$30} & \color{gray}{3.8M}\\
        \hline
        \color{gray}{Slow-8$\times$8-P \cite{slowfast}}&-& & \color{gray}{74.8} & \color{gray}{91.6} & \color{gray}{54.5$\times$30} & \color{gray}{32.5M}\\
         Slow-8$\times$8-R & - & & 74.2 & 91.3 & 54.5$\times$30 & 32.5M\\
         A3D-Slow-8$\times$8-[0.016,1] $\times$1 &-& & 75.2& 91.7&54.5$\times$30 & 32.5M \\
         A3D-Slow-8$\times$8-[0.016,1] $\times$0.5 &-& & 73.6& 90.8&20.9$\times$30 & 32.5M \\
         A3D-Slow-8$\times$8-[0.016,1] $\times$0.1 &-& & 70.9& 89.7&5.3$\times$30 & 8.3M \\
         A3D-Slow-8$\times$8-[0.016,1] $\times$0.016 &-& & 64.1& 85.0&0.9$\times$30 & 8.3M \\
         A3D-Slow-8$\times$8-[0.06,1] $\times$1 &-& & \textbf{75.6}& \textbf{91.8}& 54.5$\times$30 & 32.5M \\
         A3D-Slow-8$\times$8-[0.06,1] $\times$0.5 &-& & 74.3& 91.2& 22.5$\times$30 & 17.5M \\
         A3D-Slow-8$\times$8-[0.06,1] $\times$0.1 &-& & 70.7& 89.2&4.4$\times$30 & 13.0M \\
         \hline
        \color{gray}{SlowFast-4$\times$16-P \cite{slowfast}}&-& & \color{gray}{75.6} & \color{gray}{92.1} & \color{gray}{36.1$\times$30} & \color{gray}{34.4M}\\
        SlowFast-4$\times$16-R & - & & 74.4 & 91.5 & 36.1$\times$30 & 34.4M\\
        A3D-SlowFast-4$\times$16-[0.06,1] $\times$1 & - & & \textbf{75.7} & \textbf{92.3} & 36.1$\times$30 & 34.4M\\
        A3D-SlowFast-4$\times$16-[0.06,1] $\times$0.5 & - & & 74.9 & 91.9  & 17.4$\times$30 & 24.4M\\
        A3D-SlowFast-4$\times$16-[0.06,1] $\times$0.25 & - & & 71.6 & 90.1 & $9.0\times$30 & 24.4M\\
        \hline
        
        \hline
        
        \end{tabular}}
    \caption{Comparison of performance, computation cost (in GFLOPs $\times$ view), and parameter size of different methods.}
    \label{tab:main_result}
\end{table}

\begin{figure}[!htbp]
    \centering
    \vspace{-0.1cm}
    \includegraphics[width=0.99\linewidth]{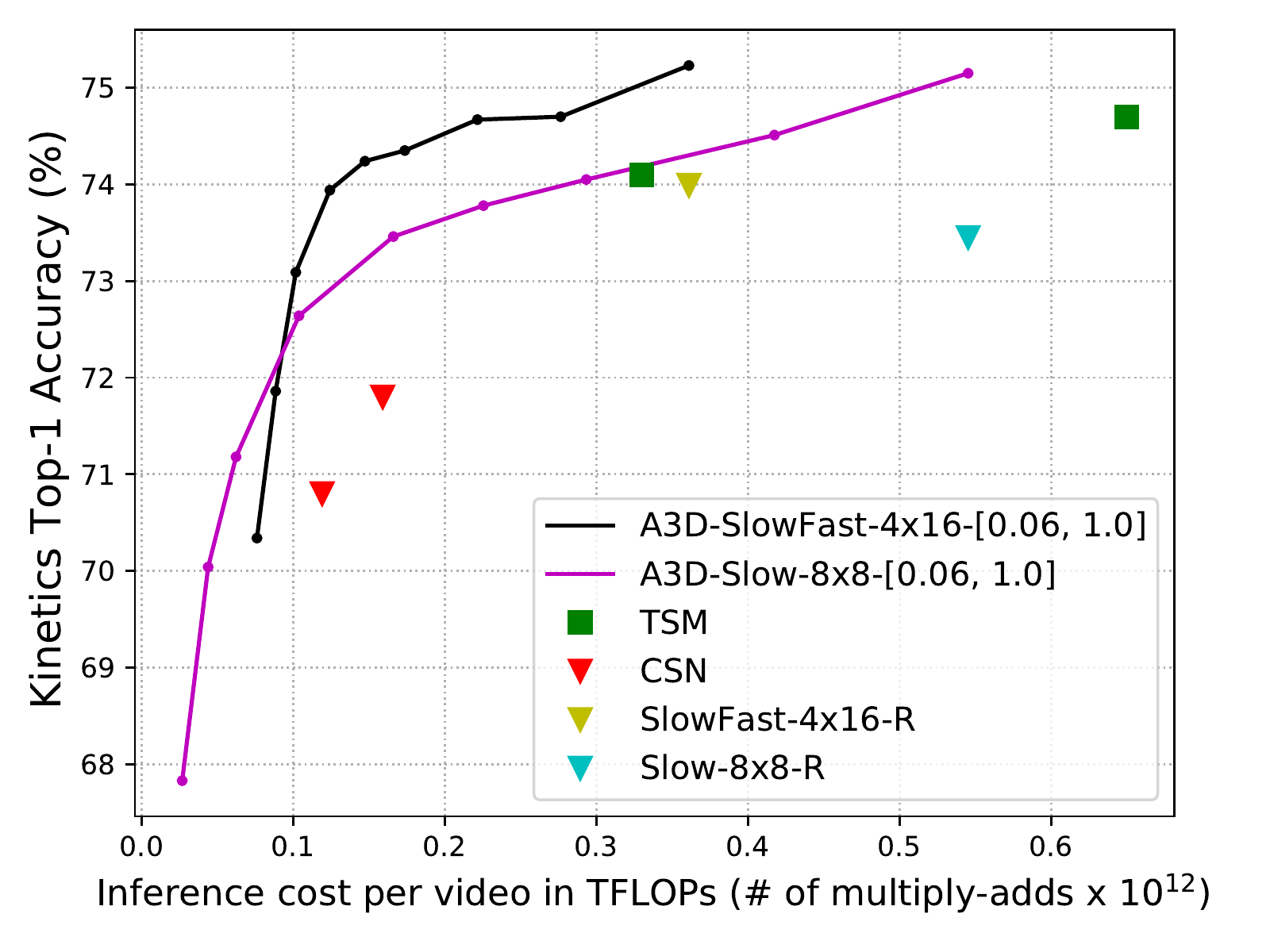}
    \vspace{-0.2cm}
    \caption{Comparison of A3D with state-of-the-art 3D networks.}
    \label{fig:10clips_all}
\end{figure}

The default testing of SlowFast uses $30$ views for each video, while some previous efficient 3D networks (\eg TSM \cite{TSM} and CSN \cite{CSN}) are based on $10$-view testing. Therefore, we show the $10$-view testing results of A3D and previous works in Fig. \ref{fig:10clips_all} for comparison. A3D-Slow-8$\times$8 and A3D-SlowFast-4$\times$16 achieve better accuracy-computation trade-off than previous methods \textbf{with one-time training}. 

\noindent\textbf{Training Cost.}
Since we sample two sub-networks in each training iteration, A3D consumes more computational cost than a single model. However, we show that the wall-clock time is only slightly longer than the full model and the total time is much shorter than independently training several models. In Table \ref{tab:trainingcost} we measure the training cost and time based on Slow-8$\times$8. We show the cost of independently training 6 models of different complexities. The training time of the full model ($\times1.0$) is measured on an 8$\times$ 2080TI GPU server with a batch-size of 64. Other sub-model times are estimated by the model complexity. The training complexity of A3D is estimated by taking the expectation of Eq. \ref{eq:2}. The training time is measured on the same server. Although the theoretical training complexity of A3D is $1.6$ times of the single full model, the wall-clock time is only slightly longer (69 vs. 58 mins/epoch) because the sub-networks share the video data of full-network in memory and do not need to decode video clips from disk again. 
Therefore, the wall-clock time of training an A3D model is much less than training several independent models.
\begin{table}[!htbp]
\huge
    \centering
    \vspace{-0.1cm}
    \resizebox{\linewidth}{!}{
    \begin{tabular}{r|c|c|c|c|c|c|c}
      & \multicolumn{6}{c|}{Slow-8$\times$8} & A3D-Slow-8$\times$8 \\
    \hline
    
    \hline
      Range & $\times$0.1 & $\times$0.3 & $\times$0.5 & $\times$0.7 & $\times$0.9 & $\times$1.0 & $\times$0.06 $\sim\times$1.0 \\
     \hline
     GFLOPs & 5.5 & 16.4 & 27.3 & 38.2 & 49.1 & 54.5 & 87.5$^{*}$ \\
     \hline
     Total & \multicolumn{6}{c|}{191} & 87.5$^{*}$ \\
     \hline
     Mins/epoch & 5.8$^{*}$ & 17.4$^{*}$ & 29.0$^{*}$ & 40.6$^{*}$ & 52.2$^{*}$ & 58 & 69 \\
     \hline
     Total & \multicolumn{6}{c|}{203} & 69 \\
    \end{tabular}
    }
    \caption{Training costs of A3D and independently training several models. $*$ indicates expected values.}
    \label{tab:trainingcost}
    \vspace{-0.1cm}
\end{table}

\noindent\textbf{Charades.}
We finetune the models trained on Kinetics-400 on Charades. For SlowFast models, we use the pre-trained models reproduced by us for a fair comparison. For A3D models, we do not perform adaptive training during finetuning. That means both SlowFast models and A3D models follow the same finetuning process on Charades. The only difference is the pre-trained models. We follow the training settings in the released codes \cite{codes}. Since we train the model on 4 GPUs, we reduce the batch-size and base learning rate by half following the linear scaling rule \cite{lrscale}. All other settings remain unchanged. As can be seen in Table \ref{tab:charades}, A3D model outperforms its counterpart (Slow-8$\times$8) by 0.9\% without increasing the computational cost. Note that the only difference lies in the pre-trained model, so the improvement demonstrate that our method helps the network learn effective and well-generalized representations which are transferable across different datasets.

\begin{table}[!htbp]
\Large
    \centering
    \vspace{-0.1cm}
    \resizebox{\linewidth}{!}{
    \begin{tabular}{l|c|c|c}
    model & pretrain & mAP & GFLOPs$\times$views \\
    \hline
    
    \hline
    CoViAR, R-50 \cite{coviar} & ImageNet & 21.9 & N/A \\
    Asyn-TF, VGG16 \cite{asyn} & ImageNet & 22.4 & N/A \\
    MultiScale TRN \cite{trn} & ImageNet & 25.2 & N/A \\
    Nonlocal, R-101 \cite{nonlocal} & ImageNet+Kinetics & 37.5 & $544 \times 30$ \\
     \hline
     Slow-8$\times$8 & Kinetics & 34.7 & $54.5 \times 30$ \\
    A3D-Slow-8$\times$8 & Kinetics & \textbf{35.6} & \textbf{$54.5 \times 30$} \\
    \end{tabular}
    }
    \caption{Comparison of different models on Charades. All models are based on R-50.}
    \label{tab:charades}
    \vspace{-0.1cm}
\end{table}

\noindent\textbf{AVA Detection.}
Similar to the experiments in Charades, we follow the same training settings as the released SlowFast codes \cite{codes}. The detector is similar to Faster R-CNN \cite{ren2015faster} with minimal modifications adopted for video. The region proposals are pre-computed by an off-the-shelf person detector. Experiments are conducted on AVA v2.1. All models (with R-50 backbone) are trained on a 4-GPU machine for 20 epochs with a batch-size of 32. The base learning rate is 0.05 with linear warm-up for the first 5 epochs. The learning rate is reduced by a factor of 10 at the 10th and 15th epochs. Both SlowFast pre-trained models and A3D pre-trained models are finetuned following the standard training procedure; the only difference is the pre-trained models. As shown in Table \ref{tab:ava}, A3D pre-trained model also outperforms SlowFast and previous methods. Note that only the pre-trained weights are different in the experiments, so the improvements are not marginal and clearly demonstrate the effectiveness of the learned representations.
\begin{table}[!htbp]
 \small
    \centering
    \vspace{-0.1cm}
    \begin{tabular}{l|c|c|c}
    model (R-50 backbone) & flow & pretrain & mAP \\ 
    \hline
    
    \hline
    I3D \cite{i3d} &  & Kinetics-400 & 14.5 \\
    I3D \cite{i3d} & \checkmark & Kinetics-400 & 15.6 \\
    ACRN, S3D \cite{actorcentric} & \checkmark & Kinetics-400 & 17.4 \\
    ATR, R-50+NL \cite{atr} & & Kinetics-400 & 20.0 \\
     \hline
     Slow-8$\times$8 & & Kinetics-400 & 20.2 \\
    A3D-Slow-8$\times$8 & & Kinetics-400 & \textbf{20.6} \\ 
    \end{tabular}
    \caption{Comparison of different models on AVA v2.1.}
    \label{tab:ava}
    \vspace{-0.2cm}
\end{table}

\subsection{Ablation Study}
\begin{figure}[!htbp]
    \centering
    \vspace{-0.2cm}
    \includegraphics[width=0.99\linewidth]{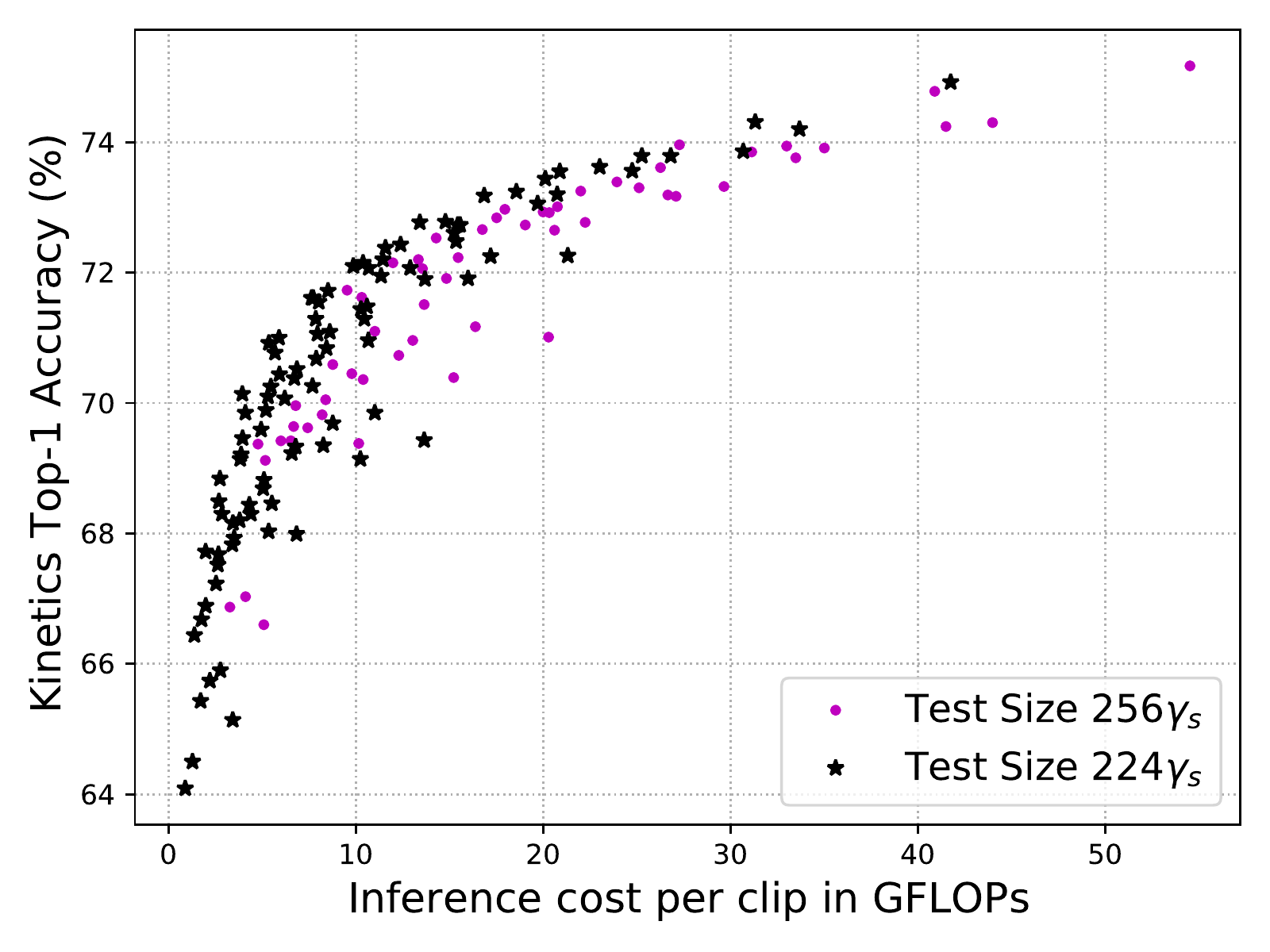}
    \vspace{-0.1cm}
    \caption{A3D-Slow-8$\times$8-[0.016, 1] testing based on crop sizes $224\gamma_{s}$ and $256\gamma_{s}$.}
    \label{fig:test_crop}
    \vspace{-0.4cm}
\end{figure}
\begin{figure}[!htbp]
    \centering
    \vspace{-0.2cm}
    \includegraphics[width=0.99\linewidth]{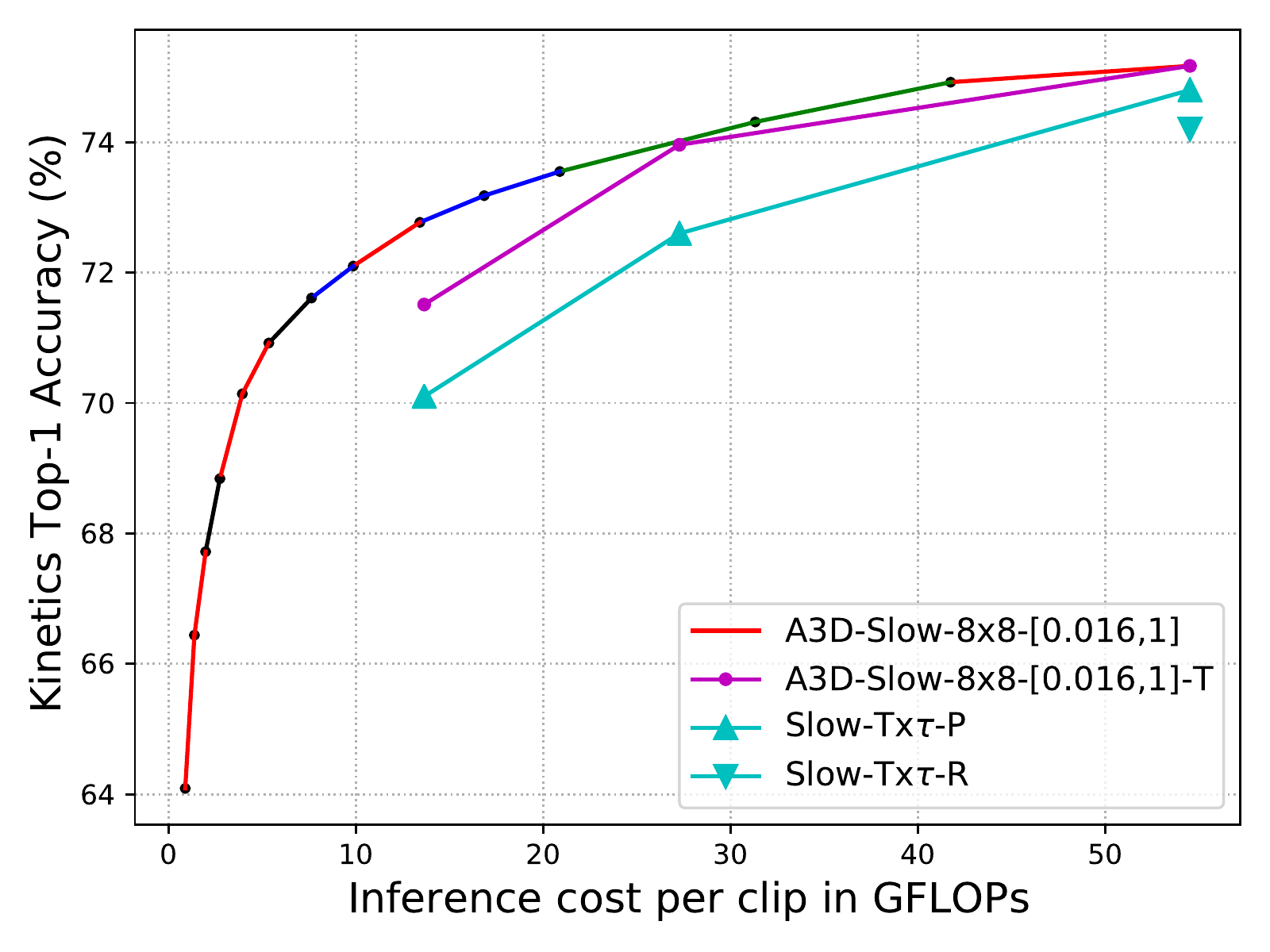}
    \vspace{-0.2cm}
    \caption{Accuracy-computation trade-off curves of A3D-Slow-8$\times$8-[0.016, 1] and Slow networks. Different line colors show different dimensions for reducing computation (\textcolor{red}{red} for spatial ($\gamma_{s}$), \textcolor{ao}{green} for temporal ($\gamma_{t}$), and \textcolor{blue}{blue} for network width ($\gamma_{w}$). \textbf{Black} indicates multiple dimensions for one trade-off step.). ``-T'' means only reducing the temporal resolution. }
    \label{fig:dimension}
    \vspace{-0.4cm}
\end{figure}

\noindent\textbf{Testing Crop Size.} The default testing crop size of SlowFast is 256, while the training crop size is 224. This is uncommon because training and testing inputs usually have the same resolution for better generalization. Although this trick improves the performance (from $73.63$ to $74.2$ for Slow-8$\times$8), the computation cost in terms of GFLOPs is also increased by $\sim$1.3$\times$. When considering accuracy-computation trade-off, this setting may not be effective, especially for the proposed A3D since it covers multiple spatial-temporal resolutions. In Fig. \ref{fig:test_crop}, we show the performances of different configurations with spatial resolutions of $256\gamma_{s}$ and $224\gamma_{s}$. The A3D-Slow-8$\times$8-[0.016,1] has a $\gamma_{s} \in \{0.57, 0.71, 0.86, 1.0\}$, thus the corresponding spatial resolutions are $\{128, 160, 192, 224\}$ for $224\gamma_{s}$ and $\{146, 182, 219, 256\}$ for $256\gamma_{s}$. Although testing with $256$ generates the best accuracy, other configurations of $224\gamma_{s}$ constantly outperform $256\gamma_{s}$ under the same computation constraints. Therefore, we only test A3D models with spatial resolution of $224\gamma_{s}$ and $256$ for all our experiments. \\
\noindent\textbf{Trade-off Dimensions.} Fig. \ref{fig:dimension} shows the accuracy-computation trade-off curves of A3D-Slow-8$\times$8-[0.016, 1] and Slow networks. ``Slow-T$\times\tau$'' means Slow networks with different temporal resolutions. ``-R'' and ``-P'' denote the reproduced results using the official codes and the numbers reported in the paper. We also provide the results of A3D only reducing temporal resolutions (``-T'') to compare with Slow networks under the same configurations. For the curve of A3D-Slow-8$\times$8-[0.016, 1], we use different colors to show different trade-off dimensions (Fig. \ref{fig:dimension}).
\begin{table}[!htbp]
\Large
    \centering
    \resizebox{\linewidth}{!}{
    \begin{tabular}{c|l|c|c|c|c}
        & model & $ S^{2} \times T,\gamma_{w}$ & top-1 & GFLOPs$\times$views& Param  \\
        \hline
        
        \hline
        \multirow{3}{0.5cm}{$\times$1}&{\color{gray} Slow-8$\times$8-P} \cite{slowfast}& \color{gray}{$256^{2}\times 8,1.0$}& \color{gray}{74.8}  & \color{gray}{54.5$\times$30} & \color{gray}{32.5M}\\
        &Slow-8$\times$8-R & $256^{2}\times 8,1.0$ & 74.2 & 54.5$\times$30 & 32.5M\\
        &A3D-Slow-8$\times$8-[0.016,1] & $256^{2}\times 8,1.0$ & \textbf{75.2}&\textbf{54.5$\times$30} & \textbf{32.5M} \\
        \hline
       \multirow{3}{0.8cm}{$\times$0.5} & \color{gray}{Slow-4$\times$16-P \cite{slowfast}} & \color{gray}{$256^{2}\times 4,1.0$} & \color{gray}{72.6} & \color{gray}{27.3$\times$30} & \color{gray}{32.5M}\\
         &A3D-Slow-8$\times$8-[0.016,1]-T  & $256^{2}\times 4,1.0$ &\textbf{74.0}&27.3$\times$30 & 32.5M \\
        
         &A3D-Slow-8$\times$8-[0.016,1]  & $224^{2}\times 6,0.9$ &73.8& \textbf{25.3$\times$30} & \textbf{26.5M} \\
         \hline
         \multirow{3}{1.2cm}{$\times$0.25}&\color{gray}{Slow-2$\times$32-P \cite{slowfast}} & \color{gray}{$256^{2}\times 2,1.0$} & \color{gray}{70.1} & \color{gray}{13.6$\times$30} & \color{gray}{32.5M}\\
          &A3D-Slow-8$\times$8-[0.016,1]-T  & $256^{2}\times 2,1.0$ &71.5&13.6$\times$30 & 32.5M \\
         &A3D-Slow-8$\times$8-[0.016,1] & $224^{2}\times 4,0.8$ &\textbf{72.8}&\textbf{13.4$\times$30} & \textbf{21.0M} \\
         
    \end{tabular}
    }
    \caption{Comparison between different trade-off strategies of A3D given the same computation constraints. }
    \label{tab:trade-off}
\end{table}
Under the same configuration, ``A3D-Slow-8$\times$8-[0.016, 1]-T'' outperforms Slow networks by a large margin. Note that the results of Slow-4$\times$16 and Slow-2$\times$32 are adopted from the paper \cite{slowfast}, and the reproduced results should be lower. Since A3D can perform trade-offs on three dimensions, ``A3D-Slow-8$\times$8-[0.016, 1]'' further surpasses ``A3D-Slow-8$\times$8-[0.016, 1]-T'' by finding better configurations. As shown in Table \ref{tab:trade-off}, ``Slow-2$\times$32'' and ``A3D-Slow-8$\times$8-[0.016, 1]-T'' use a temporal resolution of $2$ to meet $0.25\times$ computation cost, while ``A3D-Slow-8$\times$8-[0.016, 1]'' achieves $1.3\%$ higher accuracy with less GFLOPs and parameters with lower spatial resolution and network width. 
In addition, the trade-off curve of A3D brings insights about the mutual training and separated training. Based on a progressive grid-search training, X3D \cite{x3d} claims that a small network should keep a high temporal resolution, as witnessed by the configuration of X3D-M, \ie $224^{2}\times 16$ with only 3.76M parameters. While even for A3D-Slow-8$\times$8, it is more beneficial to reduce temporal resolution to $4$ before reducing network width. Therefore, the conclusions drew from separated training may not hold for a mutual training framework.\\

\begin{figure}[!htbp]
    \centering
    \vspace{-0.5cm}
    \includegraphics[width=0.98\linewidth]{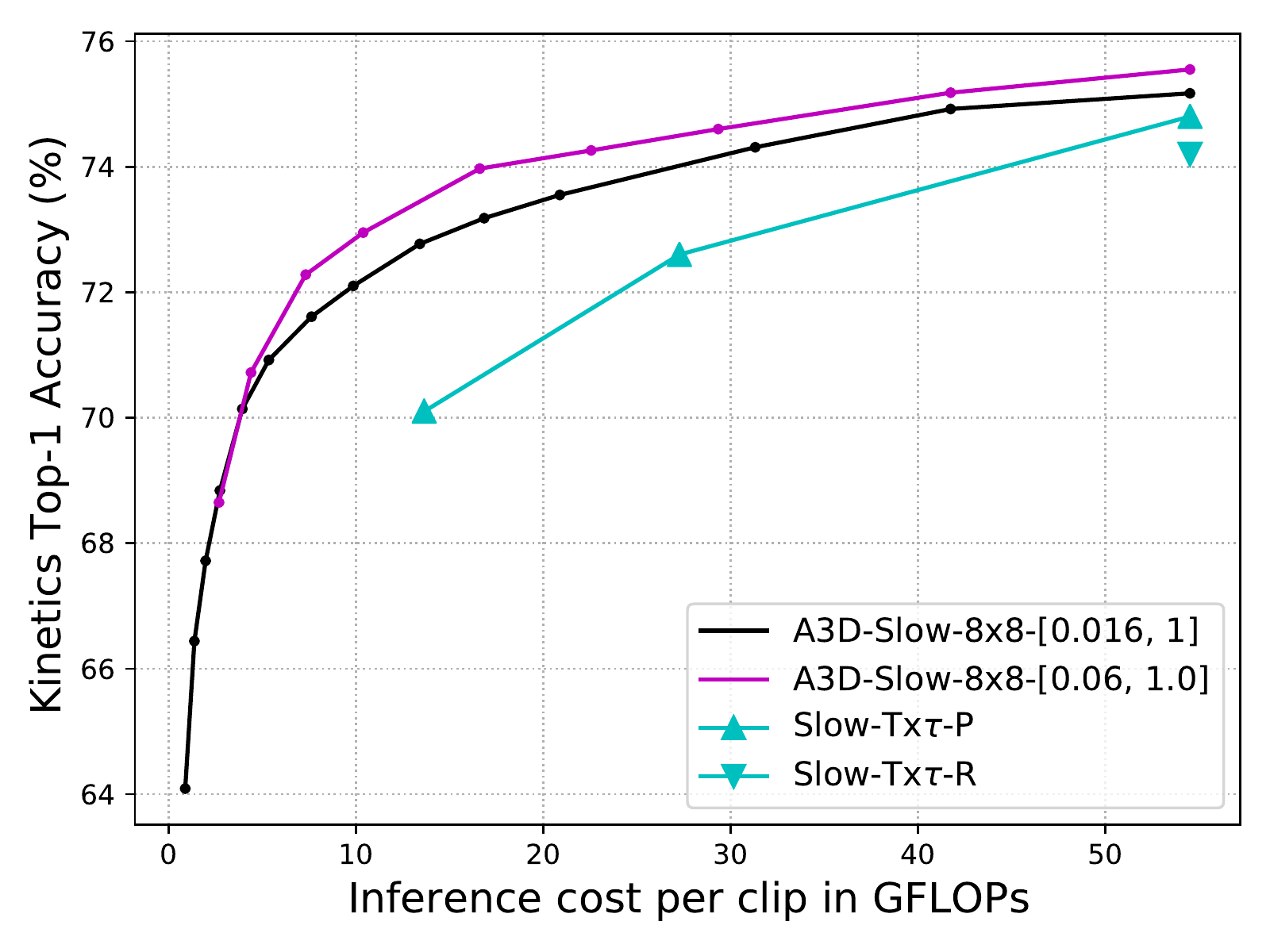}
    \vspace{-0.2cm}
    \caption{Comparison of A3D-Slow with different computational ranges.}
    \vspace{-0.2cm}
    \label{fig:coverage}
\end{figure}
\noindent\textbf{Computational Range.} To explore the effect of the computational range on performance, we compare A3D-Slow-8$\times$8-[0.016, 1] with A3D-Slow-8$\times$8-[0.06, 1] in Fig. \ref{fig:coverage}. It can be observed that a larger computational lower bound (\eg 0.06) generally provides a better accuracy-computation trade-off due to a narrower computational range. However, for the smallest FLOPs configuration ($\gamma_{w}=0.63,\gamma_{s}=0.63,\gamma_{t}=0.4$) of $[0.06,1]$, it can not trade-off between different dimensions, thus has a similar performance as $[0.016,1]$. In the \textbf{Appendix}, we present the detailed configurations of the trade-off list for A3D under these two computational ranges. 
\\ \noindent\textbf{Temporal Dimension.} To investigate the effect of the temporal dimension, we train A3D-Slow-8$\times$8-[0.06,1] without a temporal trade-off. To reduce the computation by $16\times$ (resulting in a computational range around [0.06,1]), the lower bounds of $\gamma_{s},\gamma_{t},\gamma_{w}$ are $0.63,0.4,0.63$. However, if we keep $\gamma_{t}=1$, then the lower bounds of $\gamma_{s},\gamma_{w}$ have to be $0.5,0.5$ to provide the same computational range. Table \ref{tab:temporal} shows that removing the temporal dimension in A3D leads to a lower accuracy for both the full-network ($\times 1$) and sub-networks ($\times 0.25$, $\times 0.06$).\\
\begin{table}[!htbp]
\vspace{-0.4cm}
\Large
    \centering
    \resizebox{\linewidth}{!}{
    \begin{tabular}{c|l|c|c|c|c}
        & model & $ S^{2} \times T,\gamma_{w}$ & top-1 & GFLOPs$\times$views& Param  \\
        \hline
        
        \hline
        \multirow{4}{0.8cm}{$\times$1}&{\color{gray} Slow-8$\times$8-P\cite{slowfast}}& \color{gray}{$256^{2}\times 8,1.0$}& \color{gray}{74.8}  & \color{gray}{54.5$\times$30} & \color{gray}{32.5M}\\
        &Slow-8$\times$8-R & $256^{2}\times 8,1.0$ & 74.2 & 54.5$\times$30 & 32.5M\\
        &A3D-Slow-8$\times$8-[0.06,1] w/o T & $256^{2}\times 8,1.0$ &75.1&54.5$\times$30 & 32.5M \\
        &A3D-Slow-8$\times$8-[0.06,1]  & $256^{2}\times 8,1.0$ & \textbf{75.6}&\textbf{54.5$\times$30} & \textbf{32.5M} \\
        \hline
        \multirow{3}{1.2cm}{$\times$0.25}& \color{gray}{Slow-2$\times$32-P \cite{slowfast}} & \color{gray}{$256^{2}\times 2,1.0$} & \color{gray}{70.1} & \color{gray}{13.6$\times$30} & \color{gray}{32.5M}\\
         &A3D-Slow-8$\times$8-[0.06,1] w/o T  & $224^{2}\times 8,0.6$ &73.3&15.4$\times$30 & \textbf{11.8M} \\
        
         &A3D-Slow-8$\times$8-[0.06,1]  & $224^{2}\times 5,0.73$ & \textbf{73.6}& \textbf{14.1$\times$30} & 17.5M \\
         \hline
          \multirow{2}{1.2cm}{$\times$0.06}&A3D-Slow-8$\times$8-[0.06,1] w/o T & $112^{2}\times 8,0.5$ &67.8& 2.8$\times$30 & \textbf{8.3M} \\
         &A3D-Slow-8$\times$8-[0.06,1] & $142^{2}\times 3,0.63$ &\textbf{68.7}&\textbf{2.7$\times$30} & 13.0M \\
         
    \end{tabular}
    }
    \caption{Comparison between A3D-Slow w/ and w/o temporal dimension.}
    \label{tab:temporal}
    \vspace{-0.1cm}
\end{table}

\vspace{-0.2cm}
\noindent\textbf{Mutual Training.} A3D benefits from the mutual training of different spatial-temporal resolutions, which is partly similar to multi-resolution training. To further demonstrate their difference, we train a Slow-8$\times$8 with randomly sampled spatial-temporal resolutions. The candidate resolutions are the same as those of A3D-Slow-8$\times$8-[0.06, 1]. As shown in Table \ref{tab:multiple}, simply applying spatial-temporal multi-resolution training does not improve the performance, because the network is partly trained with low resolution inputs while the testing resolution is always the highest. On the contrary, our mutual training always feeds the full-network with the highest spatial-temporal resolution, but allows sub-networks to learn multi-resolution representations. This procedure will not hurt the performance of the full-network.\\
\begin{table}[!htbp]
\small
    \centering
    \vspace{-0.4cm}
    \begin{tabular}{l|c|c}
         model& top-1 & top-5 \\
         \hline
         {\color{gray} Slow-8$\times$8-P} \cite{slowfast}& \color{gray}{74.8} & \color{gray}{91.6} \\
         Slow-8$\times$8-R & 74.2 & 91.3 \\
         Slow-8$\times$8-Multi-Resolution & \color{red}{73.4} & \color{red}{90.8} \\
         A3D-Slow-8$\times$8-[0.06,1] $\times$1 & \textbf{75.6} & \textbf{91.8}\\
    \end{tabular}
    \caption{Comparison between A3D-Slow and multi-resolution training on Slow-8$\times$8.}
    \label{tab:multiple}
    \vspace{-0.2cm}
\end{table}

\begin{figure}[!htbp]
    \centering
    \vspace{-0.2cm}
    \includegraphics[width=0.96\linewidth]{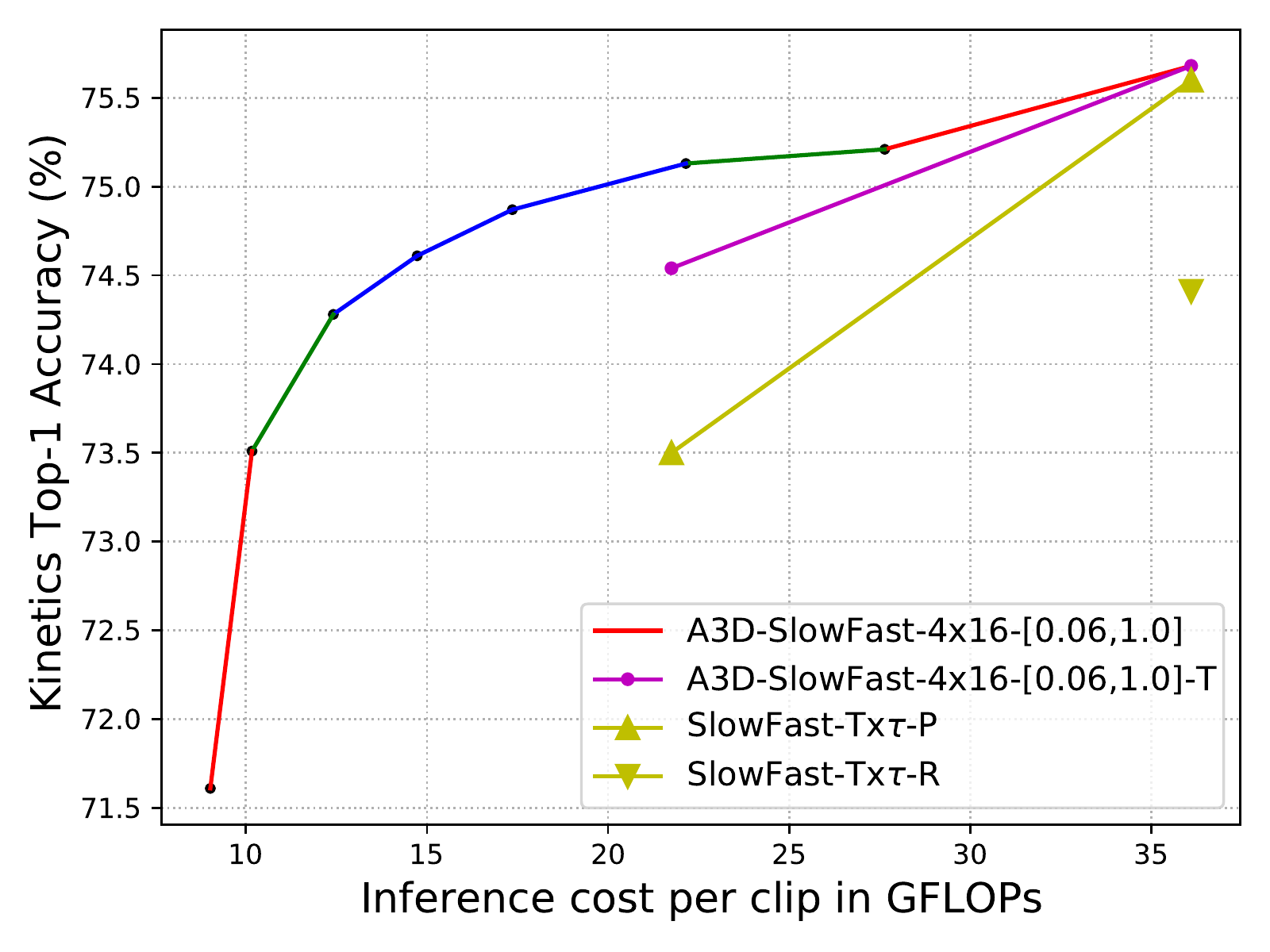}
    \vspace{-0.2cm}
    \caption{{Comparison of A3D-SlowFast and SlowFast. Different line colors show different dimensions for reducing computation (same as Fig.~\ref{fig:dimension}). 
    ``-T'' means only reducing the temporal resolution.}}
    \label{fig:A3D-Slowfast}
\end{figure}

\noindent\textbf{Multiple-pathway Trade-off.} In Fig. \ref{fig:A3D-Slowfast}, We provide the trade-off curves of A3D-SlowFast4$\times$16 and SlowFast networks (please refer to the \textbf{Appendix} for the detailed trade-off list).  Table \ref{tab:slowfast} further shows the detailed configurations under typical constraints. When the temporal resolution of Slow pathway is reduced from $4$ to $2$, the accuracy of SlowFast drops $2.1\%$, while the accuracy of A3D-SlowFast only drops $1.2\%$ due to the mutual training. With multiple-dimension trade-offs, A3D further outperforms the temporal-only (``-T'') version. 

The Fast pathway in A3D-SlowFast is not reduced for trade-offs, thus the range $[0.06,1]$ only applies for the Slow pathway.  However, the Fast pathway makes the trade-off on temporal dimension more beneficial, as the performance drop of A3D-SlowFast is only $0.1\%$ when $\gamma_{t}$ is reduced from $1.0$ to $0.75$ ($\{224^{2} \times 4\} \xrightarrow{} \{224^{2}\times 3\}$), which is much lower than that for A3D-Slow in Fig. \ref{fig:dimension}. As shown in Table \ref{tab:slowfast}, the $0.35 \times$ configuration has almost the same accuracy as the reproduced SlowFast-4$\times$16 with much less computation cost and parameters.
\begin{table}[!htbp]
\Large
    \centering
    \vspace{-0.4cm}
    \resizebox{\linewidth}{!}{
    \begin{tabular}{c|l|c|c|c|c}
        & model & $ S^{2} \times T,\gamma_{w}$ & top-1 & GFLOPs$\times$views& Param  \\
        \hline
        
        \hline
        \multirow{3}{0.8cm}{ $\times$1}& \color{gray}{SlowFast-4$\times$16-P \cite{slowfast}}& \color{gray}{$256^{2}\times 4,1.0$}& \color{gray}{75.6} & \color{gray}{36.1$\times$30} & \color{gray}{34.4M}\\
        &SlowFast-4$\times$16-R & $256^{2}\times 4,1.0$ & 74.4   & 36.1$\times$30 & 34.4M\\
        &A3D-SlowFast-4$\times$16-[0.06,1] & $256^{2}\times 4,1.0$ & \textbf{75.7}  & \textbf{36.1$\times$30} & \textbf{34.4M}\\
        \hline
        \multirow{1}{1.2cm}{$\times$0.8}&A3D-SlowFast-4$\times$16-[0.06,1]  & $ 224^{2}\times {\color{red}4},1.0$ & 75.2 & $27.6\times$30 & 34.4M \\
        \hline
        \multirow{1}{1.2cm}{$\times$0.65}&A3D-SlowFast-4$\times$16-[0.06,1]  & $ 224^{2}\times {\color{red}3},1.0$ & 75.1 & $22.1\times$30 & 34.4M \\
        \hline
        \multirow{3}{1.2cm}{$\times$0.6}& \color{gray}{SlowFast-2$\times$32-P \cite{slowfast}} & \color{gray}{$256^{2}\times 2,1.0$} & \color{gray}{73.5} & \color{gray}{21.8$\times$30} & \color{gray}{34.4M}\\
        &A3D-SlowFast-4$\times$16-[0.06,1]-T & $256^{2}\times 2,1.0$  & 74.5 & 21.8$\times$30 & 34.4M\\
        &A3D-SlowFast-4$\times$16-[0.06,1] & $224^{2}\times {\color{red} 4},0.83$  & \textbf{75.0} & \textbf{21.3$\times$30} & \textbf{24.4M}\\
        \hline
        
         \multirow{1}{1.2cm}{$\times$0.5}&A3D-SlowFast-4$\times$16-[0.06,1] & $224^{2}\times {\color{red}3},0.83$  & 74.9 & 17.4$\times$30 & 24.4M\\
         \hline
         \multirow{1}{1.2cm}{$\times$0.35}&A3D-SlowFast-4$\times$16-[0.06,1]  & $ 224^{2}\times 3,0.63$ & 74.3 & $12.4\times$30 & 14.5M \\
         \hline
         \multirow{1}{1.2cm}{$\times$0.25}&A3D-SlowFast-4$\times$16-[0.06,1]  & $ 142^{2}\times2,0.83$ & 71.6 & $9.0\times$30 &  24.4M\\
         
    \end{tabular}
    }
    \caption{Comparison between A3D-SlowFast-4$\times$16 and SlowFast under different computation constraints.
    }
    \label{tab:slowfast}
    \vspace{-0.3cm}
\end{table}

\section{Conclusion}
This paper presents the first adaptive 3D network (A3D) which can achieve adaptive accuracy-efficiency trade-offs at runtime for video action recognition with a single model. It considers network width and input spatio-temporal resolution and randomly samples different spatial-temporal-width configurations in network training. A spatial-temporal distillation scheme is developed to facilitate knowledge transfer among different configurations. The training paradigm is generic and applicable to any 3D networks. Using the same backbone, A3D outperforms the state-of-the-art SlowFast \cite{slowfast} networks under various computation constraints on Kinetics. Extensive evaluations also validate the effectiveness of the learned representations of A3D for cross dataset and task transfer. 

{\small
\bibliographystyle{ieee_fullname}
\bibliography{egbib}
}


\appendix
\section{Appendix}

\subsection{Full Trade-off List}
\label{sec:trade-off-list}
\paragraph{A3D-Slow-8$\times$8-[0.016, 1].} We show the detailed configurations of the trade-off list for A3D-Slow-8$\times$8-[0.016, 1] in Table \ref{tab:trade-off-list-slow-0.016}. The configurations correspond to the black curve in Fig. \ref{fig:coverage} of the main paper. 

\begin{table}[!htbp]
\Large
    \centering
    \resizebox{\linewidth}{!}{
    \begin{tabular}{c|c|c|c|c}
        $\rho$ & $ S^{2} \times T,\gamma_{w}$ & top-1 & GFLOPs$\times$views& Param  \\
        \hline
        
        \hline
        $1.00\times$ & $256^{2}\times 8,1.0$ & 75.2 & 54.5$\times$30 & 32.5M\\
        $0.77\times$ & $224^{2}\times 8,1.0$ & 74.9 & 41.7$\times$30 & 32.5M\\
        $0.57\times$ & $224^{2}\times 6,1.0$ & 74.3 & 31.3$\times$30 & 32.5M\\
        $0.38\times$ & $224^{2}\times 4,1.0$ & 74.2 & 20.9$\times$30 & 32.5M\\
        $0.31\times$ & $224^{2}\times 4,0.9$ & 73.2 & 16.8$\times$30 & 26.5M\\
        $0.25\times$ & $224^{2}\times 4,0.8$ & 72.8 & 13.4$\times$30 & 21.0M\\
        $0.18\times$ & $192^{2}\times 4,0.8$ & 72.1 & 9.8$\times$30 & 21.0M\\
        $0.14\times$ & $192^{2}\times 4,0.7$ & 71.6 & 7.6$\times$30 & 16.0M\\
        $0.10\times$ & $224^{2}\times 4,0.5$ & 70.9 & 5.3$\times$30 & 8.3M\\
        $0.07\times$ & $192^{2}\times 4,0.5$ & 70.1 & 3.9$\times$30 & 8.3M\\
        $0.05\times$ & $160^{2}\times 4,0.5$ & 68.8 & 2.7$\times$30 & 8.3M\\
        $0.04\times$ & $192^{2}\times 2,0.5$ & 67.7 & 2.0$\times$30 & 8.3M\\
        $0.03\times$ & $160^{2}\times 2,0.5$ & 66.4 & 1.4$\times$30 & 8.3M\\
        $0.016\times$ & $128^{2}\times 2,0.5$ & 64.1 & 0.9$\times$30 & 8.3M\\
         
    \end{tabular}
    }
    \caption{Full trade-off list of A3D-Slow-8$\times$8-[0.016, 1]. }
    \label{tab:trade-off-list-slow-0.016}
\end{table}

\paragraph{A3D-Slow-8$\times$8-[0.06, 1].} We show the detailed configurations of the trade-off list for A3D-Slow-8$\times$8-[0.06, 1] in Table \ref{tab:trade-off-list-slow-0.06}. The configurations correspond to the curve (``\textcolor{byzantine}{\bf --}") in Fig. \ref{fig:coverage} of the main paper. 

\begin{table}[!htbp]
\Large
    \centering
    \resizebox{\linewidth}{!}{
    \begin{tabular}{c|c|c|c|c}
        $\rho$ & $ S^{2} \times T,\gamma_{w}$ & top-1 & GFLOPs$\times$views& Param  \\
        \hline
        
        \hline
        $1.00\times$ & $256^{2}\times 8,1.0$ & 75.6 & 54.5$\times$30 & 32.5M\\
        $0.77\times$ & $224^{2}\times 8,1.0$ & 75.2 & 41.7$\times$30 & 32.5M\\
        $0.54\times$ & $224^{2}\times 8,0.83$ & 74.6 & 29.3$\times$30 & 22.5M\\
        $0.41\times$ & $224^{2}\times 8,0.73$ & 74.3 & 22.5$\times$30 & 17.5M\\
        $0.30\times$ & $224^{2}\times 8,0.63$ & 74.0 & 16.6$\times$30 & 13.0M\\
        $0.19\times$ & $224^{2}\times 5,0.63$ & 73.0 & 10.4$\times$30 & 13.0M\\
        $0.13\times$ & $178^{2}\times 5,0.63$ & 72.3 & 7.3$\times$30 & 13.0M\\
        $0.08\times$ & $178^{2}\times 3,0.63$ & 70.7 & 4.4$\times$30 & 13.0M\\
        $0.05\times$ & $142^{2}\times 3,0.63$ & 68.7 & 2.7$\times$30 & 13.0M\\
         
    \end{tabular}
    }
    \caption{Full trade-off list of A3D-Slow-8$\times$8-[0.06, 1]. }
    \label{tab:trade-off-list-slow-0.06}
\end{table}

\paragraph{A3D-SlowFast-4$\times$16-[0.06, 1].} We show the detailed configurations of the trade-off list for A3D-SlowFast-4$\times$16-[0.06, 1] in Table \ref{tab:trade-off-list-slowfast-0.06}. The configurations correspond to the curve in Fig. \ref{fig:A3D-Slowfast} of the main paper. 

\begin{table}[!htbp]
\Large
    \centering
    \resizebox{\linewidth}{!}{
    \begin{tabular}{c|c|c|c|c}
        $\rho$ & $ S^{2} \times T,\gamma_{w}$ & top-1 & GFLOPs$\times$views& Param  \\
        \hline
        
        \hline
        $1.00\times$ & $256^{2}\times 4,1.0$ & 75.7 & 36.1$\times$30 & 34.5M\\
        $0.76\times$ & $224^{2}\times 4,1.0$ & 75.2 & 27.6$\times$30 & 34.5M\\
        $0.61\times$ & $224^{2}\times 3,1.0$ & 75.1 & 22.1$\times$30 & 34.5M\\
        $0.48\times$ & $224^{2}\times 3,0.83$ & 74.9 & 17.4$\times$30 & 24.4M\\
        $0.41\times$ & $224^{2}\times 3,0.73$ & 74.6 & 14.7$\times$30 & 19.2M\\
        $0.34\times$ & $224^{2}\times 3,0.63$ & 74.3 & 12.4$\times$30 & 14.5M\\
        $0.28\times$ & $224^{2}\times 2,0.63$ & 73.5 & 10.2$\times$30 & 14.5M\\
        $0.24\times$ & $178^{2}\times 2,0.63$ & 72.4 & 8.8$\times$30 & 14.5M\\
        $0.23\times$ & $142^{2}\times 2,0.73$ & 71.3 & 8.3$\times$30 & 19.2M\\
        $0.21\times$ & $142^{2}\times 2,0.63$ & 71.0 & 7.6$\times$30 & 14.5M\\
         
    \end{tabular}
    }
    \caption{Full trade-off list of A3D-SlowFast-4$\times$16-[0.06, 1].}
    \label{tab:trade-off-list-slowfast-0.06}
\end{table}

\subsection{Full Trade-off Table} 
\paragraph{A3D-Slow-8$\times$8-[0.016, 1].} In Table \ref{tab:trade-off-table-slow-0.016}, we show the results of all the configurations tested in our paper for A3D-Slow-8$\times$8-[0.016, 1]. The green color highlights the configurations that are selected as the trade-off list in Sec. \ref{sec:trade-off-list}.

\begin{table}[!htbp]
\Large
    \centering
    \resizebox{\linewidth}{!}{
    \begin{tabular}{l|c|c|c|c|c|c}
        \diagbox{$S^{2}\times T$}{$\gamma_{w}$} & 1.0 & 0.9 & 0.8 & 0.7 & 0.6 & 0.5 \\
        \hline
        
        \hline
        $256^{2}\times 8$ & \cellcolor{green} 75.2 & - & - & - &  - & - \\
        $224^{2}\times 8$ & \cellcolor{green} 74.9 & 74.2 & 73.8 & 73.2 & 72.7 & 72.1\\
        $224^{2}\times 6$ & \cellcolor{green} 74.3 & 73.8 & 73.4 & 72.7 & 72.4 & 71.6\\
        $224^{2}\times 4$ & \cellcolor{green} 73.6 & \cellcolor{green} 73.2 & \cellcolor{green} 72.8 & 72.2 & 71.6 & \cellcolor{green} 70.9\\
        $224^{2}\times 2$ & 71.3 & 70.8 & 70.4 & 69.9 & 69.2 & 68.5\\
        $192^{2}\times 8$ & 73.9 & 73.6 & 73.1 & 72.6 & 72.0 & 71.3\\
        $192^{2}\times 6$ & 73.6 & 73.2 & 72.8 & 72.2 & 71.7 & 71.0\\
        $192^{2}\times 4$ & 72.5 & 72.4 & \cellcolor{green} 72.1 & \cellcolor{green} 71.6 & 70.8 & \cellcolor{green} 70.1\\
        $192^{2}\times 2$ & 70.3 & 70.1 & 69.6 & 69.1 & 68.3 & \cellcolor{green} 67.7\\
        $160^{2}\times 8$ & 72.3 & 72.3 & 71.9 & 71.5 & 70.7 & 70.3\\
        $160^{2}\times 6$ & 71.9 & 72.1 & 71.4 & 71.1 & 70.4 & 69.9\\
        $160^{2}\times 4$ & 71.0 & 71.1 & 70.5 & 70.1 & 69.5 & \cellcolor{green} 68.8\\
        $160^{2}\times 2$ & 68.0 & 68.4 & 68.2 & 67.7 & 66.9 & \cellcolor{green} 66.4\\
        $128^{2}\times 8$ & 69.4 & 69.9 & 69.7 & 69.3 & 68.7 & 67.9\\
        $128^{2}\times 6$ & 69.1 & 69.4 & 69.2 & 68.8 & 68.2 & 67.5\\
        $128^{2}\times 4$ & 68.0 & 68.5 & 68.0 & 67.8 & 67.2 & 66.7\\
        $128^{2}\times 2$ & 65.1 & 65.9 & 65.1 & 65.4 & 64.5 & \cellcolor{green} 64.1\\
         
    \end{tabular}
    }
    \caption{Full trade-off table of A3D-Slow-8$\times$8-[0.016, 1]. }
    \label{tab:trade-off-table-slow-0.016}
\end{table}

\paragraph{A3D-Slow-8$\times$8-[0.06, 1].} In Table \ref{tab:trade-off-table-slow-0.06}, we show the results of all the configurations tested in our paper for A3D-Slow-8$\times$8-[0.06, 1].  The green color highlights the configurations that are selected as the trade-off list in Sec. \ref{sec:trade-off-list}.

\paragraph{A3D-SlowFast-4$\times$16-[0.06, 1].} In Table \ref{tab:trade-off-table-slowfast-0.06}, we show the results of all the configurations tested in our paper for A3D-SlowFast-4$\times$16-[0.06, 1].  The green color highlights the configurations that are selected as the trade-off list in Sec. \ref{sec:trade-off-list}.

\begin{table}[!htbp]
\Large
    \centering
    \resizebox{\linewidth}{!}{
    \begin{tabular}{l|c|c|c|c|c}
        \diagbox{$S^{2}\times T$}{$\gamma_{w}$} & 1.0 & 0.93 & 0.83 & 0.73 & 0.63 \\
        \hline
        
        \hline
        $256^{2}\times 8$ & \cellcolor{green}75.6 & - & - & - &  - \\
        $224^{2}\times 8$ & \cellcolor{green} 75.2 & 74.6 & \cellcolor{green} 74.6 & \cellcolor{green} 74.3 & \cellcolor{green} 74.0 \\
        $224^{2}\times 5$ & 74.2 & 73.8 & 73.8 & 73.6 & \cellcolor{green} 73.0 \\
        $224^{2}\times 3$ & 72.6 & 72.3 & 72.2 & 71.9 & 71.7 \\
        $178^{2}\times 8$ & 73.7 & 73.6 & 73.5 & 73.2 & 73.0 \\
        $178^{2}\times 5$ & 73.0 & 72.8 & 72.7 & 72.4 & \cellcolor{green} 72.3 \\
        $178^{2}\times 3$ & 71.0 & 71.2 & 71.2 & 71.0 & \cellcolor{green} 70.7 \\
        $142^{2}\times 8$ & 71.5 & 71.7 & 71.8 & 71.5 & 71.3 \\
        $142^{2}\times 5$ & 70.7 & 71.1 & 71.0 & 70.7 & 70.5 \\
        $142^{2}\times 3$ & 68.9 & 69.4 & 69.3 & 68.9 & \cellcolor{green} 68.7 \\
         
    \end{tabular}
    }
    \caption{Full trade-off table of A3D-Slow-8$\times$8-[0.06, 1]. }
    \label{tab:trade-off-table-slow-0.06}
\end{table}

\begin{table}[!htbp]
\Large
    \centering
    \resizebox{\linewidth}{!}{
    \begin{tabular}{l|c|c|c|c|c}
        \diagbox{$S^{2}\times T$}{$\gamma_{w}$} & 1.0 & 0.93 & 0.83 & 0.73 & 0.63 \\
        \hline
        
        \hline
        $256^{2}\times 4$ & \cellcolor{green}75.7 & - & - & - &  - \\
        $224^{2}\times 4$ & \cellcolor{green} 75.2 & 75.1 & 75.0 & 74.7 & 74.6 \\
        $224^{2}\times 3$ & \cellcolor{green} 75.1 & 75.0 & \cellcolor{green} 74.9 & \cellcolor{green} 74.6 & \cellcolor{green} 74.3 \\
        $224^{2}\times 2$ & 74.4 & 74.2 & 74.1 & 73.9 & \cellcolor{green} 73.5 \\
        $178^{2}\times 4$ & 74.2 & 74.3 & 74.1 & 73.7 & 73.4 \\
        $178^{2}\times 3$ & 73.9 & 74.0 & 73.9 & 73.7 & 73.3 \\
        $178^{2}\times 2$ & 73.1 & 73.2 & 73.1 & 72.9 & \cellcolor{green} 72.4 \\
        $142^{2}\times 4$ & 72.4 & 72.6 & 72.6 & 72.6 & 72.0 \\
        $142^{2}\times 3$ & 72.1 & 72.5 & 72.4 & 72.3 & 71.8 \\
        $142^{2}\times 2$ & 71.2 & 71.7 & 71.6 & \cellcolor{green} 71.3 & \cellcolor{green} 71.0 \\
         
    \end{tabular}
    }
    \caption{Full trade-off table of A3D-SlowFast-4$\times$16-[0.06, 1]. }
    \label{tab:trade-off-table-slowfast-0.06}
\end{table}

\end{document}